\UseRawInputEncoding
\documentclass[runningheads,a4paper]{llncs}

\usepackage{amssymb,amsmath}
\usepackage{subfigure}
\usepackage{graphicx}
\usepackage{environ}
\usepackage{url}
\usepackage{stmaryrd}

\usepackage{url}
\urldef{\mailsa}\path|tdenoeux@utc.fr|
\newcommand{\keywords}[1]{\par\addvspace\baselineskip
\noindent\keywordname\enspace\ignorespaces#1}

\newcommand{\reels}{\mathbb{R}}

\newcommand{\esp}{\mathbb{E}}

\def\Unif{\textsf{Unif}}

\newcommand{\height}{\textsf{hgt}}

\newcommand{\calL}{{\cal L}}

\newcommand{\calT}{{\cal T}}

\newcommand{\tN}{{\widetilde{N}}}

\newcommand{\tX}{{\widetilde{X}}}

\newcommand{\tTheta}{{\widetilde{\Theta}}}

\newcommand{\tY}{{\widetilde{Y}}}

\def\bbeta{{\boldsymbol{\beta}}}

\def\btheta{\boldsymbol{\theta}}

\def\bx{{\boldsymbol{x}}}

\def\bw{{\boldsymbol{w}}}

\def\Finf{{\underline{F}}}
\def\Fsup{{\overline{F}}}

\def\cut#1#2{{}^#1#2}
\newcommand{\fracpar}[2]{\left(\frac{#1}{#2}\right)}

\def\GFN{\textsf{GFN}}

\newcommand{\bi}{\begin{itemize}}
\newcommand{\ei}{\end{itemize}}
\newcommand{\be}{\begin{enumerate}}
\newcommand{\ee}{\end{enumerate}}
\newcommand{\bd}{\begin{description}}
\newcommand{\ed}{\end{description}}

\NewEnviron{eqs}[1]
{
\begin{subequations}
\label{#1}
\begin{align}
\BODY
\end{align}
\end{subequations}
}

\begin{document}

\mainmatter  

\title{An Evidential  Neural Network Model for Regression Based on Random Fuzzy Numbers
 }

\titlerunning{Evidential neural network model for regression}

%
%

\author{Thierry Den{\oe}ux\orcidID{0000-0002-0660-5436} }
\authorrunning{T. Den{\oe}ux}

\institute{
Universit\'e de technologie de Compi\`egne, CNRS, \\
UMR 7253 Heudiasyc, France \\
Institut universitaire de France, Paris, France\\
\mailsa }

%
%

\toctitle{Lecture Notes in Computer Science}
\tocauthor{Authors' Instructions}
\maketitle

\begin{abstract}
We introduce a  distance-based neural network model for regression, in which prediction uncertainty is quantified by a belief function on the real line. The model interprets the distances of the input vector to prototypes as pieces of evidence represented by Gaussian random fuzzy numbers (GRFN's) and combined by the generalized product intersection rule, an operator that extends Dempster's rule to random fuzzy sets. The network output is a GRFN that can be summarized by three numbers characterizing the most plausible predicted value, variability around this value, and epistemic uncertainty. Experiments with real datasets  demonstrate the very good performance of the method as compared to state-of-the-art evidential and statistical learning algorithms.
\keywords{Evidence theory, Dempster-Shafer theory, belief functions, machine learning, random fuzzy sets.}
\end{abstract}

\section{Introduction}
\label{sec:intro}

The Dempster-Shafer (DS) theory of evidence is a general mathematical framework for reasoning and making decisions based on imprecise and uncertain information \cite{shafer76}\cite{denoeux20b}. This framework is based on the representation of independent pieces of evidence by belief functions, and on their combination by a conjunctive operator called Dempster's rule. The greater number of degrees of freedom offered by belief functions, as compared to probabilities,  makes it possible to distinguish between two situations of high uncertainty: equally supported hypotheses on the one hand, and total lack of support on the other hand, the latter situation characterizing complete ignorance.

In machine learning, DS theory has been mainly applied to classification and clustering tasks, in which the set of elementary hypotheses (or frame of discernment) is finite. In particular, several methods have been proposed to learn \emph{evidential classifiers}, i.e., classifiers representing  prediction uncertainty  by belief functions. In the first such classifier, the evidential $K$-nearest neighbor (EKNN) rule \cite{denoeux95a},  each neighbor of a feature vector to be classified is represented by a simple mass function, and the mass functions from the $K$ nearest neighbors are combined by Dempster's rule. The evidential neural network (ENN) introduced in \cite{denoeux00a} is based on the same principle, the elementary mass functions being computed based on distances to prototypes. Recently, this distance-based approach has been extended to deep networks \cite{tong21b}\cite{tong21a}\cite{huang22}, by adding a DS layer to a deep architecture; output mass functions are then computed based on distances to prototypes in the space of high-level features extracted by convolutional layers.  

Applying DS theory to regression is more challenging, because in regression tasks the  frame of discernment is typically the real line or a real interval, whereas most tools of DS theory have been developed for finite frames. This difficulty can be circumvented by discretizing the response variable, as proposed in \cite{denoeux97c}, in which a neural network model for regression directly extending the  ENN model was introduced. The output of this model is a mass function with disjoint intervals and the whole frame as focal sets. Another approach, introduced in \cite{petit99a}\cite{petit04}, is to modify the EKNN rule by combining simple mass functions focussed either on a single real number, or on a fuzzy number in the case of learning data with fuzzy response variable. The output mass function then has a finite number of crisp or fuzzy focal sets. This method, called EVREG, was shown in \cite{petit04} to yield good results in the case of crisp data, and to efficiently handle uncertain response data (such as provided by an unreliable sensor). However, the $K$ nearest neighbor approach breaks down as dimension grows, and it cannot compete with state-of-the-art regression methods. 

In this paper, we propose another evidential neural network model for regression inspired from the ENN model. This new model, called ENNreg, uses the formalism of Gaussian random fuzzy numbers (GRFN's) recently introduced in \cite{denoeux22}. A GRFN is a random fuzzy subset of the real line, which can be described as a Gaussian possibility distribution whose mode is a Gaussian random variable.  GRFN's induce belief functions and can be combined using a generalization of Dempster's rule. In ENNreg,  GRFN's associated to each of the prototypes are combined to yield an output GRFN quantifying prediction uncertainty.

The rest of this paper is organized as follows. The general framework of epistemic random fuzzy sets and the  GRFN model are first recalled in Section \ref{sec:RFS}. The proposed ENNreg model is then introduced in Section \ref{sec:model}. Experimental results are reported in Section \ref{sec:exper}, and Section \ref{sec:concl} concludes the paper.

\section{Epistemic Random Fuzzy Sets}
\label{sec:RFS}

The theory of epistemic random fuzzy sets (ERFS) was introduced in \cite{denoeux21a} and \cite{denoeux22} as a general framework encompassing both DS theory and possibility theory. We first recall some important definitions in Section \ref{subsec:Gen}.  Gaussian random fuzzy numbers, a parametric family of ERFS's on the real line are then  described in Section \ref{subsec:GRFN}.

\subsection{General Framework}
\label{subsec:Gen}

Let $(\Omega,\Sigma_\Omega,P)$ be a probability space and let $(\Theta,\Sigma_\Theta)$ be a measurable space. Let $\tX$ be a mapping from $\Omega$ to the set $[0,1]^\Theta$ of fuzzy subsets of $\Theta$. For any $\alpha\in [0,1]$, let $\cut{\alpha}{\tX}$ be the mapping from $\Omega$ to $2^\Theta$ defined as
\[
\cut{\alpha}{\tX}(\omega)=\cut{\alpha}{[\tX(\omega)]},
\]
where $\cut{\alpha}{[\tX(\omega)]}$  is the weak $\alpha$-cut of $\tX(\omega)$.  If for any $\alpha\in [0,1]$, $\cut{\alpha}{\tX}$ is $\Sigma_\Omega-\Sigma_\Theta$ strongly measurable \cite{nguyen78},  the tuple  $(\Omega,\Sigma_\Omega,P,\Theta, \Sigma_\Theta,\tX)$  is said to be a \emph{random fuzzy set} (also called a \emph{fuzzy random variable})  \cite{couso11}. When there is no possible confusion about the domain and codomain, we will refer to mapping $\tX$ itself as a random fuzzy set. 

 In ERFS theory,  random fuzzy sets represent unreliable and fuzzy evidence. In this model, we see $\Omega$ as a \emph{set of interpretations} of a piece of evidence about a variable $\btheta$ taking values in $\Theta$. If interpretation $\omega\in \Omega$ holds, we know that ``$\btheta \text{ is } \tX(\omega)$'', i.e., $\btheta$  is constrained by the possibility distribution defined by $\tX(\omega)$. We qualify such random fuzzy sets as \emph{epistemic}, because they encode a state of knowledge about some variable $\btheta$. If all images $\tX(\omega)$ are crisp, then $\tX$ defines an ordinary random set. If mapping $\tX$ is constant, then it is equivalent to specifying a unique fuzzy subset of $\Theta$, which defines a possibility distribution.

\paragraph{Belief and plausibility functions.} In the following, we will assume  that random fuzzy set $\tX$ is   \emph{normalized}, i.e.,   that it verifies the following conditions: (1) For all $\omega\in\Omega$, $\tX(\omega)$ is either the empty set, or a normal fuzzy set, i.e., $\height(\tX(\omega))=\sup_{\theta\in \Theta} \tX(\omega)(\theta)\in\{0,1\}$; (2)  $P(\{\omega \in \Omega: \tX(\omega)=\emptyset\})=0$. For any $\omega\in\Omega$, let $\Pi_\tX(\cdot\mid \omega)$ be the possibility measure on $\Theta$ induced by $\tX(\omega)$:
\begin{equation}
\label{eq:defPi}
\Pi_\tX(B\mid \omega)=\sup_{\theta\in B} \tX(\omega)(\theta),
\end{equation}
and let  $N_\tX(\cdot\mid \omega)$ be the dual necessity measure:
\[
N_\tX(B\mid \omega)=\begin{cases}
1-\Pi_\tX(B^c\mid \omega) & \text{if } \tX(\omega)\neq \emptyset\\
0 &  \text{otherwise, }
\end{cases}
\]
where $B^c$ denotes the complement of $B$. The mappings   $Bel_\tX$ and $Pl_\tX$ from $\Sigma_\Theta$ to $[0,1]$ defined as
\begin{equation}
\label{eq:defBel}
Bel_\tX(B)=\int_\Omega  N_\tX(B\mid \omega) dP(\omega) 
\end{equation}
and
\begin{equation}
\label{eq:defPl}
Pl_\tX(B)=\int_\Omega  \Pi_\tX(B\mid \omega) dP(\omega) =1-Bel_\tX(B^c)
\end{equation}
are, respectively,   belief  and  plausibility functions.

\paragraph{Combination.} Consider two  epistemic random fuzzy sets $(\Omega_i,\Sigma_i,P_i,\Theta, \Sigma_\Theta,\tX_i)$, $i=1,2$, encoding independent pieces of evidence. The independence assumption means here that the relevant probability measure on the joint measurable space $(\Omega_1\times\Omega_2,\Sigma_1\otimes\Sigma_2$) is the product measure $P_1\times P_2$. If  interpretations $\omega_1\in \Omega_1$ and $\omega_2\in \Omega_2$ both hold, we know that ``$\btheta \text{ is } \tX_1(\omega_1)$'' and ``$\btheta \text{ is } \tX_2(\omega_2)$''. It is then natural to combine the fuzzy sets $\tX_1(\omega_1)$ and $\tX_2(\omega_2)$ by an intersection operator. As discussed in \cite{denoeux21a}, the normalized product intersection operator $\varodot$ is  suitable for combining fuzzy information from independent sources and it is associative. We  thus consider the mapping $\tX_\varodot(\omega_1,\omega_2)=\tX_1(\omega_1)\varodot \tX_2(\omega_2)$, assumed to be $\Sigma_1\otimes\Sigma_2$-$\Sigma_\Theta$ strongly measurable.

If $\height(\tX_1(\omega_1)\tX_2(\omega_2))=0$, the two interpretations $\omega_1$ and $\omega_2$ are inconsistent and they must be discarded. If  $\height(\tX_1(\omega_1)\tX_2(\omega_2))=1$, the two interpretations are fully consistent. If $0<\height(\tX_1(\omega_1)\tX_2(\omega_2))<1$, $\omega_1$ and $\omega_2$ are \emph{partially consistent}. The \emph{soft normalization} proposed in \cite{denoeux22} consists in conditioning the product probability $P_1\times P_2$ by the fuzzy subset $\tTheta^*$ of consistent pairs of interpretations, with membership function $\tTheta^*(\omega_1,\omega_2)= \height\left(\tX_1(\omega_1)\cdot \tX_2(\omega_2)\right)$. Alternatively, we can use a \emph{hard normalization} operation, which consists in conditioning $P_1\times P_2$ by the crisp set $\Theta^*$ of interpretations that are not fully inconsistent, described formally as 
\[
\Theta^*=\{(\omega_1,\omega_2)\in\Omega_1\times\Omega_2:  \height(\tX_1(\omega_1)\tX_2(\omega_2))>0\}.
\]
Both combination rules, with soft or hard normalization, are commutative and associative, and both of them generalize Dempster's rule. In the following, we will use hard normalization as it leads to simpler calculations. This operation will be referred to as the \emph{generalized product-intersection rule with hard normalization}, and the corresponding operator will be denoted by $\boxplus$.

\subsection{Gaussian Random Fuzzy Numbers}
\label{subsec:GRFN}

A \emph{Gaussian Fuzzy Number} (GFN) is a fuzzy subset of $\reels$ with membership function
\[
\varphi(x;m,h)=\exp\left(-\frac{h}{2}(x-m)^2\right),
\]
where  $m\in\reels$ is the mode and  $h\in [0,+\infty]$ is the precision. Such a fuzzy number will be denoted by $\GFN(m,h)$. The normalized product intersection of two GFN's  $\GFN(m_1,h_1)$ and  $\GFN(m_2,h_2)$ is a GFN 
$\GFN(m_{12},h_{12})=\GFN(m_1,h_1)\varodot \GFN(m_2,h_2)$, with $m_{12}=(h_1 m_1+h_2 m_2)/(h_1+h_2)$ and $h_{12}=h_1+h_2$.

Let $(\Omega,\Sigma_\Omega,P)$ be a probability space and let $M: \Omega\rightarrow\reels$ be a Gaussian random variable (GRV) with mean $\mu$ and variance $\sigma^2$. The random fuzzy set  $\tX:\Omega\rightarrow [0,1]^\reels$ defined as 
\[
\tX(\omega) = \GFN(M(\omega),h)
\]
is called a \emph{Gaussian random fuzzy number} (GRFN) with mean $\mu$, variance $\sigma^2$ and precision $h$, which we write   $\tX\sim\tN(\mu,\sigma^2, h)$. A GRFN is, thus, defined by a location parameter $\mu$, and two parameters $h$ and $\sigma^2$ corresponding, respectively, to possibilistic and probabilistic uncertainty. 

A GRFN can be seen either as a generalized GRV with fuzzy mean, or as a generalized GFN with random mode. In particular, a GRFN $\tX$ with infinite precision $h=+\infty$ is  equivalent to a GRV with mean $\mu$ and variance $\sigma^2$, which we can write: $\tN(\mu,\sigma^2,+\infty)=N( \mu,\sigma^2)$. If $\sigma^2=0$,  $M$ is a constant random variable taking value $\mu$, and $\tX$ is a possibilistic variable with possibility distribution $\GFN(\mu,h)$. Another case of interest is that where $h=0$, in which case $\tX(\omega)(x)=1$ for all $\omega\in\Omega$ and each $x\in\reels$, and the belief function induced by $\tX$ is vacuous. 

As shown in \cite{denoeux22}, the  plausibility and belief of any real interval $[x,y]$ are given by the following formulas:
\begin{subequations}
\label{eq:BelPl}
\begin{multline}
Pl_\tX([x,y])=\Phi\fracpar{y-\mu}{\sigma} -\Phi\fracpar{x-\mu}{\sigma} + \\pl_\tX(x)\Phi\fracpar{x-\mu}{\sigma\sqrt{h\sigma^2+1}} + 
pl_\tX(y)\left[1-\Phi\fracpar{y-\mu}{\sigma\sqrt{h\sigma^2+1}}\right],
\end{multline}
and
\begin{multline}
Bel_\tX([x,y])=Pl_\tX([x,y])-
pl_\tX(x)\Phi\fracpar{(x+y)/2-\mu}{\sigma\sqrt{h\sigma^2+1}} - \\
pl_\tX(y)\left[1-\Phi\fracpar{(x+y)/2-\mu}{\sigma\sqrt{h\sigma^2+1}}\right],
\end{multline}
where $\Phi$ is the standard normal cumulative distribution function (cdf), and
\begin{equation}
pl_\tX(x)=\frac{1}{\sqrt{1+h\sigma^2}}\exp\left(- \frac{h(x-\mu)^2}{2(1+h\sigma^2)}\right)
\end{equation}
\end{subequations}
is the contour function. Denoting by $\Finf_\tX(x)=Bel_\tX((-\infty,x])$ and $\Fsup_\tX(x)=Pl_\tX((-\infty,x])$, respectively, the lower and upper cdf's of $\tX$, its lower and upper expectations  are
\[
\esp_*(\tX)=\int_{-\infty}^{+\infty} x \, d\Fsup_\tX(x) =\mu-\sqrt{\frac{\pi}{2h}} 
\]
and
\[
\esp^*(\tX)=\int_{-\infty}^{+\infty} x \, d\Finf_\tX(x)=\mu+\sqrt{\frac{\pi}{2h}}.
\]

The usefulness of GRFN's as a model of uncertain information about a real quantity arises from the fact that GRFN's can easily be combined by the generalized product-intersection rule, with soft or hard normalization \cite{denoeux22}. Here, we only consider hard normalization, which is used in the proposed regression model described in Section \ref{sec:model}. Given two GRFN's $\tX_1\sim\tN(\mu_1,\sigma_1^2, h_1)$ and $\tX_2\sim\tN(\mu_2,\sigma_2^2, h_2)$, we have $\tX_1\boxplus\tX_2 \sim\tN(\mu_{12},\sigma_{12}^2, h_{12})$, with
\[
\mu_{12}=\frac{h_1 \mu_1+h_2 \mu_2}{h_1+h_2}, \quad \sigma_{12}^2=\frac{h_1^2 \sigma_1^2+h_2^2 \sigma_2^2}{(h_1+h_2)^2}, \quad \text{and} \quad  h_{12}=h_1+h_2.
\]

\section{Neural network model}
\label{sec:model}

The ENN (Evidential Neural Network) model introduced in \cite{denoeux00a} for classification is based on prototypes in input space, each one having degrees of membership to the different classes. In this model, each prototype provides a piece of evidence regarding the class of a test instance. This evidence is represented by a DS mass function defined from the class membership degrees of the prototype and the distance from the input vector. The mass functions induced by the prototypes are then combined by Dempster's rule. Here, we propose a similar model for regression, called ENNreg. The propagation equations and the loss function are given, respectively,  in Sections \ref{subsec:propag} and \ref{subsec:loss}.

\subsection{Propagation Equations}
\label{subsec:propag}
 We consider $J$ prototypes $\bw_j  \in \reels^p$, $j=1,\ldots,J$, where $p$ is the dimension of the input space. The activation of prototype $j$ for input $\bx$ is
\[
a_j(\bx)=\exp(-\gamma_j^2 \|\bx-\bw_j\|^2),
\]
where $\gamma_j$ is a positive scale parameter. The evidence of prototype $j$ is represented by a GRFN  
$
\tY_j(\bx)\sim\tN(\mu_j(\bx),\sigma_j^2,a_j(\bx)h_j),
$ 
where $\sigma_j^2$ and $h_j$ are variance and precision parameters for prototype $j$; the mean   $\mu_j(\bx)$ is defined as 
$
\mu_j(\bx)=\bbeta_j^T\bx+\alpha_j,
$ 
where $\bbeta_j$ is a $p$-dimensional vector of coefficients, and $\alpha_j$ is a scalar parameter. The vector $\psi_j$ of parameters associated to prototype $j$ is, thus, $\psi_j=(\bw_j,\gamma_j,\bbeta_j,\alpha_j,\sigma_j^2,h_j)$.

The output $\tY(\bx)$ for input $\bx$ is computed by combining the GRFN's $\tY_j(\bx)$, $j=1,\ldots,J$ induced by the $J$ prototypes using the $\boxplus$ operator. It is a GRFN $\tY(\bx)\sim\tN(\mu(\bx),\sigma^2(\bx),h(\bx))$, with 
\[
\mu(\bx)=\frac{\sum_{j=1}^J a_j(\bx)h_j \mu_j(\bx)}{\sum_{k=1}^J a_k(\bx)h_k}, \quad 
\sigma^2(\bx)=\frac{\sum_{j=1}^J a^2_j(\bx)h^2_j \sigma_j^2}{\left(\sum_{k=1}^J a_k(\bx)h_k\right)^2},
\]
and $h(\bx)=\sum_{j=1}^J a_j(\bx) h_j$. Some special cases are of interest:
\be
\item If $\bbeta_j=0$ for all $j$, then $\mu_j(\bx)=\alpha_j$, and $\mu(\bx)$ is identical to the output of a radial basis function (RBF) neural network with hidden-to-output weights $h_j\alpha_j$ and normalized outputs;
\item If $J=1$ and $\gamma_1=0$, $\mu(\bx)=\bbeta_1^T\bx+\alpha_1$, $\sigma^2(\bx)=h_1\sigma_1^2$ and $h(\bx)=h_1$. We then have a linear model with constant variance.
\ee

\subsection{Loss Function}
\label{subsec:loss}

We want to fit the model described in the previous section in such a way that the observed values of the response variable have a high degree of belief and a high plausibility. Because the degree of belief is zero for a single real value, we consider a small interval $[y-\epsilon,y+\epsilon]$, denoted as $[y]_\epsilon$, around each observed value $y$, and we define  the following  loss function:
\begin{equation}
\label{eq:loss1}
\calL_{\lambda,\epsilon}(y,\tY)=-\lambda \ln Bel_\tY([y]_\epsilon) - (1-\lambda) \ln Pl_\tY([y]_\epsilon),
\end{equation}
where  $y$ is the true response, $\tY\sim\tN(\mu,\sigma^2,h)$, and $\lambda \in [0,1]$ is a hyperparameter. This loss function is minimal for a perfect forecast, such that $\mu=y$, $h=+\infty$ and $\sigma^2\rightarrow 0$. With a fixed variance $\sigma^2$, the term $Bel_\tY([y]_\epsilon)$ is maximized for $\mu=y$ and $h=+\infty$, while the term  $Pl_\tY([y]_\epsilon)$ is maximized for $h=0$, whatever $\mu$. Hyperparameter $\lambda$ thus determines the precision of the predictions. 
We can also remark that, when $\epsilon$ is small, we have, for a probabilistic GRFN $\tY\sim\tN(\mu,\sigma^2,+\infty)$, $\calL_{\lambda,\epsilon}(y,\tY)\approx - \ln  \phi(\frac{y-\mu}\sigma) -\ln\epsilon$, where $\phi$ denotes the standard normal probability density function; loss function \eqref{eq:loss1} then becomes equivalent to  minus the log-likelihood.

Using a training set $\calT=\{(\bx_1,y_1),\ldots,(\bx_n,y_n)\}$, we minimize the regularized average loss
\[
C(\Psi)= \frac1n\sum_{i=1}^n \calL_{\lambda,\epsilon}(y_i,\tY(\bx_i)) + \frac{\xi}J \sum_{j=1}^J h_j,
\]
where $\Psi=(\psi_1,\ldots,\psi_J)$ is the vector of all parameters, and $\xi$ is a regularization coefficient. We note that setting $h_j=0$ amounts to removing prototype $j$, as the GRFN $\tY_j(\bx)$ becomes vacuous for any input $\bx$. Increasing  $\xi$ results in more cautious predictions and a more parsimonious model.

\section{Experimental results}
\label{sec:exper}

We  first give an illustrative example in Section \ref{subsec:ex}. Results from a comparative experiment are then  reported in Section \ref{subsec:compar}.

\subsection{Illustrative example}
\label{subsec:ex}

\begin{figure}
\centering  
\includegraphics[width=0.7\textwidth]{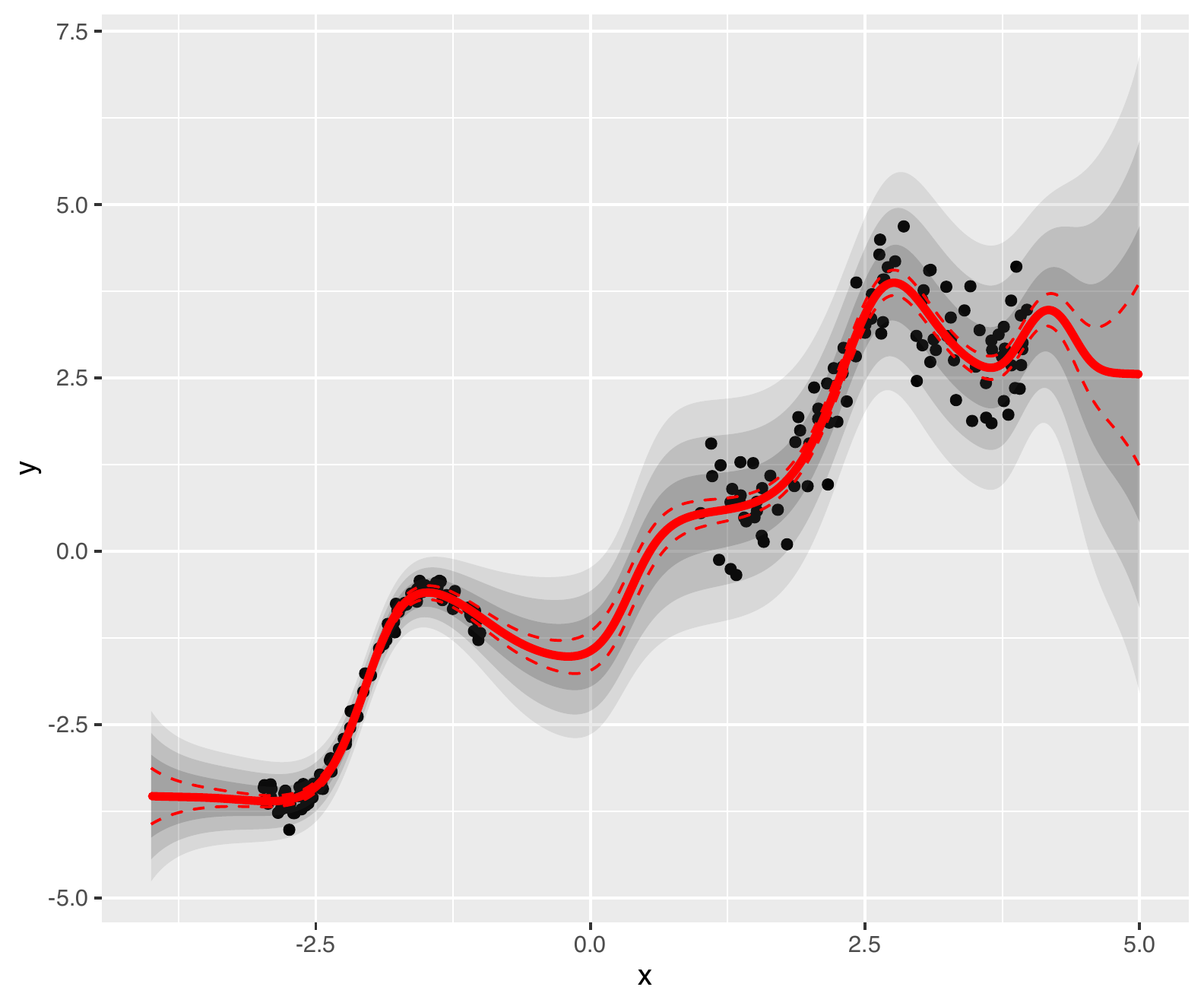}
\caption{Learning data and predictions for the illustrative example. The red solid and broken lines correspond, respectively, to the expected values $\mu(x)$, and to the  lower and upper expectations $\esp_*(\tY(x))$, $\esp^*(\tY(x))$. Prediction intervals at levels $1-\alpha\in\{0.5,0.9,0.99\}$ are shown as grey areas. \label{fig:ex}}
\end{figure}

As an illustrative example, we consider data with $p=1$ input from the following distribution:
\be
\item The input $X$ is drawn from a mixture of two uniform distributions on $[-3,-1]$ and $[1,4]$: $X\sim 0.5 \, \Unif(-3,-1) + 0.5 \, \Unif(1,4)$;
\item Given $X=x$, $Y=x+\sin 3x + \eta$, where $\eta$ is a Gaussian random variable with zero mean and variance $\sigma^2=0.01$ if $x<0$ and $\sigma^2=0.3$ otherwise.
\ee
We generated a learning set of size $n=200$ and a test set of size $n_t=1000$ from that distribution. We trained a network with $J=10$ prototypes initialized by the $k$-means algorithm, with $\lambda=0.95$, $\xi=10^{-3}$ and $\epsilon=0.01$. Figure \ref{fig:ex} shows the expected values $\mu(x)$, together with the lower and upper expectations $\esp_*(\tY(x))$, $\esp^*(\tY(x))$, as well as prediction intervals of the form
\begin{equation}
\label{eq:Finfsup}
[\Fsup^{-1}_{\tY(x)}(\alpha/2), \Finf^{-1}_{\tY(x)}(1-\alpha/2)],
\end{equation}
for $1-\alpha\in\{0.5,0.9,0.99\}$. The estimated coverage probabilities of these intervals are, respectively, 0.76, 0.96 and 0.997, which suggests that expression \eqref{eq:Finfsup} provides conservative prediction intervals. As can be seen from Figure \ref{fig:ex}, the prediction intervals are wider when the variance of the data is larger, and in regions of the input space where there is no data.

\subsection{Comparative experiment}
\label{subsec:compar}

The performance of ENNreg model was compared to those of six alternative regression methods on four  datasets from the UCI Machine Learning Repository\footnote{Available at \url{https://archive.ics.uci.edu/ml/}.}. The methods are:
\bi
\item The two evidential regression algorithms published so far: the neural network model introduced in \cite{denoeux97c} (referred to as ENN97) and EVREG \cite{petit04}:
\item Three state-of-the-art nonlinear regression algorithms  with Radial Basis Function Kernel: Relevance Vector Machines (RVM), Support Vector Machines  (SVM), and Gaussian Process  (GP);
\item The Random Forest (RF) algorithm, which is often considered as one of the best statistical learning procedures.
\ei
For all methods, except ENN97 and EVREG, we used the implementation in the R package {\tt caret} \cite{kuhn21}. Each dataset was split randomly into a training set and a test set containing, respectively, 2/3 and 1/3 of the observations. All predictors were scaled to have zero mean and unit standard deviation. For each method,  hyperparameters were tuned by 10-fold cross-validation. For ENN97, the number $M$ of classes was set to 10. For ENNreg, we used $J=30$, $\lambda=0.9$ and $\epsilon=0.01\sigma_y$ (where $\sigma_y$ is the standard deviation of the response variable) for all the simulations; only $\xi$ was tuned by cross-validation. 

The results are reported in Table \ref{tab:results}. We can see that ENNreg performs much better than the two other evidential methods on all datasets, and also better that the state-of-the-art methods for most datasets (it was only outperformed by RF on the Concrete dataset). From these results, it appears that ENNreg  not only provides informative outputs with uncertainty quantification, but is also very competitive in terms of prediction accuracy.

\begin{table}
\caption{Test mean squared errors of ENNreg and six alternative algorithms on four UCI datasets. (See the description of the methods in the text). \label{tab:results}} 
\begin{center}
\begin{tabular}{lcc|ccccccc}
\hline
            &$n$& $p$ & ENNreg & ENN97 & EVREG & RVM & SVM & GP &  RF\\
            \hline
Boston &506  &13& 8.72 &15.78 & 19.82 &14.86  &9.74 &17.10 &10.68\\
Energy &768  &9& 0.342& 5.303& 4.266& 0.721 &0.440 &2.324 &0.462\\
Concrete &1,030&8& 28.32 &62.0 &71.4 &42.3 &29.6 &52.7 &25.6\\
Yacht & 308 & 6 & 0.462 & 6.662& 42.045 & 2.771 & 3.295& 33.721&  0.908\\
\hline
\end{tabular}
\end{center}
\end{table}

\section{Conclusions}
\label{sec:concl}

The evidential  distance-based neural network  described in this paper can be seen as regression counterpart of the ENN model introduced in \cite{denoeux00a} for classification. Both models are based on prototypes, and interpret the distances of the input vector to the prototypes as pieces of evidence. In ENN,  pieces of evidence are represented by mass functions on the finite frame of discernment and are combined by Dempster's rule. In ENNreg, the frame of discernment is the real line; pieces of evidence are represented by GRFN's and combined by the generalized product-intersection rule with hard normalization, which generalizes Dempster's rule to random fuzzy sets. 

The output of ENNreg for input vector $\bx$ is a GRFN defined by three numbers: a point prediction $\mu(\bx)$, a variance $\sigma^2(\bx)$ measuring random uncertainty, and a precision $h(\bx)$, which can be seen as representing epistemic uncertainty. Experimental results show that the method outperforms previous evidential regression models in terms of mean squared error, and that it also performs better than or as well as some of the state-of-the-art nonlinear regression models. In future work, we will further investigate the calibration properties of the output belief functions, and study their potential to faithfully represent prediction uncertainty, particularly in information fusion contexts. We will also compare our approach to that of Cella and Martin \cite{cella21b}, who propose a method, applicable to any regression algorithm, for constructing a  predictive possibility distribution with some well-defined frequentist properties.


\begin{thebibliography}{10}

\bibitem{cella21b}
L.~Cella and R.~Martin.
\newblock Valid inferential models for prediction in supervised learning
  problems.
\newblock {\em Researchers.One}, 2021.
\newblock \url{https://researchers.one/articles/21.12.00002v2}.

\bibitem{couso11}
I.~Couso and L.~S\'anchez.
\newblock Upper and lower probabilities induced by a fuzzy random variable.
\newblock {\em Fuzzy Sets and Systems}, 165(1):1--23, 2011.

\bibitem{denoeux95a}
T.~Den{\oe}ux.
\newblock A $k$-nearest neighbor classification rule based on {
  Dempster-Shafer} theory.
\newblock {\em IEEE Trans. on Systems, Man and Cybernetics}, 25(05):804--813,
  1995.

\bibitem{denoeux97c}
T.~Den{\oe}ux.
\newblock Function approximation in the framework of evidence theory: A
  connectionist approach.
\newblock In {\em Proceedings of the 1997 International Conference on Neural
  Networks (ICNN'97)}, volume~1, pages 199--203, Houston, June 1997. IEEE.

\bibitem{denoeux00a}
T.~Den{\oe}ux.
\newblock A neural network classifier based on {Dempster-Shafer} theory.
\newblock {\em IEEE Trans. on Systems, Man and Cybernetics A}, 30(2):131--150,
  2000.

\bibitem{denoeux21a}
T.~Den{\oe}ux.
\newblock Belief functions induced by random fuzzy sets: A general framework
  for representing uncertain and fuzzy evidence.
\newblock {\em Fuzzy Sets and Systems}, 424:63--91, 2021.

\bibitem{denoeux20b}
T.~Den{\oe}ux, D.~Dubois, and H.~Prade.
\newblock Representations of uncertainty in artificial intelligence: Beyond
  probability and possibility.
\newblock In P.~Marquis, O.~Papini, and H.~Prade, editors, {\em A Guided Tour
  of Artificial Intelligence Research}, volume~1, chapter~4, pages 119--150.
  Springer Verlag, 2020.

\bibitem{denoeux22}
T.~Den{\oe}"ux.
\newblock Reasoning with fuzzy and uncertain evidence using epistemic random
  fuzzy sets: General framework and practical models.
\newblock {\em Fuzzy Sets and Systems}, 2022.
\newblock \url{https://doi.org/10.1016/j.fss.2022.06.004}.

\bibitem{huang22}
L.~Huang, S.~Ruan, P.~Decazes, and T.~Den{\oe}ux.
\newblock Lymphoma segmentation from {3D PET-CT} images using a deep evidential
  network, 2022.
\newblock arXiv preprint 2201.13078, \url{https://arxiv.org/abs/2201.13078}.

\bibitem{kuhn21}
M.~Kuhn.
\newblock {\em caret: Classification and Regression Training}, 2021.
\newblock R package version 6.0-90,
  \url{https://CRAN.R-project.org/package=caret}.

\bibitem{nguyen78}
H.~T. Nguyen.
\newblock On random sets and belief functions.
\newblock {\em Journal of Mathematical Analysis and Applications}, 65:531--542,
  1978.

\bibitem{petit99a}
S.~Petit-Renaud and T.~Den{\oe}ux.
\newblock Handling different forms of uncertainty in regression analysis: a
  fuzzy belief structure approach.
\newblock In A.~Hunter and S.~Pearsons, editors, {\em Symbolic and quantitative
  approaches to reasoning and uncertainty (ECSQARU'99)}, pages 340--351,
  London, June 1999. Springer Verlag.

\bibitem{petit04}
S.~Petit-Renaud and T.~Den{\oe}ux.
\newblock Nonparametric regression analysis of uncertain and imprecise data
  using belief functions.
\newblock {\em International Journal of Approximate Reasoning}, 35(1):1--28,
  2004.

\bibitem{shafer76}
G.~Shafer.
\newblock {\em A mathematical theory of evidence}.
\newblock Princeton University Press, Princeton, N.J., 1976.

\bibitem{tong21b}
Z.~Tong, P.~Xu, and T.~Den{\oe}ux.
\newblock An evidential classifier based on {Dempster-Shafer} theory and deep
  learning.
\newblock {\em Neurocomputing}, 450:275--293, 2021.

\bibitem{tong21a}
Z.~Tong, P.~Xu, and T.~Den{\oe}"ux.
\newblock Evidential fully convolutional network for semantic segmentation.
\newblock {\em Applied Intelligence}, 51:6376--6399, 2021.

\end{thebibliography}

\end{document}